\documentclass{llncs}

\usepackage{hyperref, verbatim, graphicx}

\begin{document}
\title{Dataflow Matrix Machines as a Generalization of Recurrent Neural Networks}

\author{Michael Bukatin\inst{1}\and Steve Matthews\inst{2}\ and Andrey Radul\inst{3}}

\institute{HERE North America LLC\\
Burlington, Massachusetts, USA\\ 
\email{bukatin@cs.brandeis.edu}
\and
Department of Computer Science\\
University of Warwick\\
Coventry, UK\\
\email{Steve.Matthews@warwick.ac.uk}
\and
Project Fluid\\
Cambridge, Massachusetts, USA\\ 
\email{aor021@gmail.com}}

\maketitle

\begin{abstract}Dataflow matrix machines are a powerful generalization of recurrent neural networks.
They work with multiple types of arbitrary linear streams, multiple types of powerful neurons,
and allow to incorporate higher-order constructions. We expect them to be useful in
machine learning and probabilistic programming, and in the synthesis
of dynamic systems and of deterministic and probabilistic programs. 
\end{abstract}

\section{Introduction}
Dataflow matrix machines~\cite{BukatinMatthewsDMM} can be understood as recurrent neural networks (RNNs) generalized as follows.
Neurons can have different types. Neurons are not limited to receiving and emitting streams of
real numbers, but can receive and emit streams of other vectors (depending on the type of
a neuron).

Among possible types of vectors are probability distributions and signed measures
over a wide class of spaces (including spaces of discrete objects), and samples from the
underlying spaces can be passed over the links  as the representations of the distributions
and measures in question.

Thus it is possible to send streams of discrete objects overs the links of so generalized
neural nets while retaining the capabilities to take meaningful linear combinations of
those streams (Section~\ref{sec:linear_streams}).

Because built-in neurons can be pretty powerful, dataflow matrix machines of very small
size can already exhibit complex dynamics. So meaningful dataflow matrix machines can
be quite compact, which is typical also for probabilistic programs and less typical
for  conventional RNNs.

As a powerful new tool, dataflow matrix machines can be used in various ways. Here we would
like to focus on the aspects related to program synthesis (also known as program learning, symbolic
regression, etc).

\subsection{Linear Models of Computation and RNNs}

One way which might improve the performance of program learning systems would be to find
continuous models of computations allowing for continuous deformations of software.
Computational architectures which admit the notion of linear combination of execution runs are
particularly attractive in this sense and allow to express regulation of gene expression in
the context of genetic programming~\cite{BukatinMatthewsLinear}. It turns out that if one computes with
linear streams, such as probabilistic sampling and generalized animations, one can
parametrize large classes of programs by matrices of real numbers~\cite{BukatinMatthewsDMM},
obtaining {\em dataflow matrix machines}.

In particular, recurrent neural networks provide examples of such classes of
programs, where linear streams are taken to be streams of real numbers with
element-wise arithmetic, and there is usually a very limited number of types of
non-linear transformations of those streams associated with neurons (it is often
the case that all neurons are of the same type).

Turing universality of some classes of recurrent neural networks and first schemas to compile conventional programming
languages into recurrent neural networks became known at least 20 years ago 
(see~\cite{JNetoSiegelmannCosta} and references therein). Nevertheless, 
recurrent neural networks are not used as a general purpose programming platform,
or even as a platform to program recurrent neural networks themselves. The reason is that universality
is not enough, one also needs practical convenience and power of available primitives.

\subsection{Dataflow Matrix Machines as a Programming Platform}

Dataflow programming languages, including languages oriented towards
work with streams of continuous data (e.g. LabVIEW, Pure Data), found some degree of
a more general programming use within their application domains.

Dataflow matrix machines have architecture which generalizes both recurrent neural networks
and the core architecture of dataflow languages working with the streams of continuous data.

Dataflow matrix machines allow to include neurons encoding
primitives which transform the networks (as long as it makes sense for
those primitives to be parts of linear combinations), and thus
potentially allow to move towards using generalized RNNs
to program generalized RNNs in a higher-order fashion.

This gives hope that, on one hand, one would be able to use methods proven successful
in learning the topology and the weights of recurrent neural networks to synthesize
dataflow matrix machines, and that at the same time one would be able to use
dataflow flow matrix machines as a software engineering platform for various
purposes, including the design, transformation, and learning of recurrent neural
networks and dataflow matrix machines themselves.

\section{Linear Streams}\label{sec:linear_streams}

Two prominent and highly expressive classes of linear streams are probabilistic sampling
and generalized animations~\cite{BukatinMatthewsLinear}. 

The linear combinations for each type of linear stream might be implemented on the level of vectors, 
which is what we do for the streams of real numbers in the
recurrent neural networks, and for the streams of generalized images in generalized animations.

However, the linear combination might be also implemented only on the level of streams,
which allows to represent large or infinite vectors by their compact representatives. For example, probability distributions can be represented
by samples from those distributions, and linear combinations with positive coefficients can be implemented as stochastic remixes of the respective
streams of samples~\cite{BukatinMatthewsLinear}. This provides opportunities to consider the architectures which are hybrid
between probabilistic programming and RNNs.

\section{Dataflow Matrix Machines}

Here we briefly describe the formal aspect of dataflow matrix machines. We follow~\cite{BukatinMatthewsDMM},
but transpose the matrix notation. We fix a particular
{\em signature} by taking a finite number of neuron types, each with its own fixed finite non-negative arity (zero arity
corresponds to inputs) and associated nonlinear transform.

We take a countable number of copies of neurons for each neuron type from the signature. Then we
have a countable set of inputs of those operations, $Y_i$, and a countable set of their
outputs, $X_j$.
Associate with each $Y_i$ a linear combination of all $X_j$ with real coefficients $a_{ij}$.
We require that no more than finite number of elements of the matrix $(a_{ij})$ are nonzero.

Thus we have a countable-sized program, namely a countable dataflow graph, all but
a finite part of which is suppressed by zero coefficients. Any finite dataflow graph
over a particular signature can be embedded into a universal countable dataflow graph
over this signature in this fashion.

Hence we represent programs over a fixed signature as countable-sized real-valued matrices with
no more than finite number of nonzero elements, and any program evolution would be
a trajectory in this space of matrices.

\section{Further Work: Learning High-Level Readable Programs}

There is quite a bit of interest recently in using recurrent neural networks and
related machines to learn algorithms (see e.g.~\cite{SReeddeFreitas} and references
therein). However, a typical result of program learning is a program which is
functional, but is almost impossible to read and comprehend. At the same time,
there is a lot of interest and progress in learning readable structures in various areas,
and in understanding the details of learned models. For example, in recent years
people demonstrated progress in automatic generation of readable mathematical
proofs, in automatically capturing syntactic patterns of software written by humans,
and in understanding and visualization the details of functioning of learned neural
models~\cite{MGanesalingamGowers,AKarpathy}.

One of the longer-term objectives here is to be able to automatically learn
higher level programs which not just work,
but are also readable and amenable to understanding by people.\footnote{{\bf May 2018 note:} Reference paper on dataflow matrix machines (DMMs) research in 2016-2017 is
\url{https://arxiv.org/abs/1712.07447}}

\end{document}